%% file: main.tex
\documentclass[runningheads]{llncs}

\usepackage{eccv}

\usepackage{graphicx}
\usepackage{booktabs}
\usepackage{rotating}
\usepackage{xspace}
\usepackage{soul}

\usepackage[accsupp]{axessibility}  

\usepackage[pagebackref,breaklinks,colorlinks,citecolor=eccvblue]{hyperref}

\usepackage{orcidlink}

\usepackage[dvipsnames]{xcolor}
\definecolor{mygreen}{rgb}{0.201, 0.802, 0.201}

\newcommand{\method}[0]{{\fontfamily{txtt}\selectfont{CosHand}}\xspace}
\newcommand{\hoee}[0]{hand\xspace}

\begin{document}

\title{Author Guidelines for ECCV Submission} 
\title{Controlling the World by Sleight of Hand} 


\author{Sruthi Sudhakar \and Ruoshi Liu \and Basile Van Hoorick \and Carl Vondrick \and Richard Zemel}
\institute{Columbia University \\
{\href{https://coshand.cs.columbia.edu/}{{\url{coshand.cs.columbia.edu}}}}
}
\authorrunning{Sudhakar et al.}


\maketitle
\begin{center}
    \includegraphics[width=0.98\linewidth]{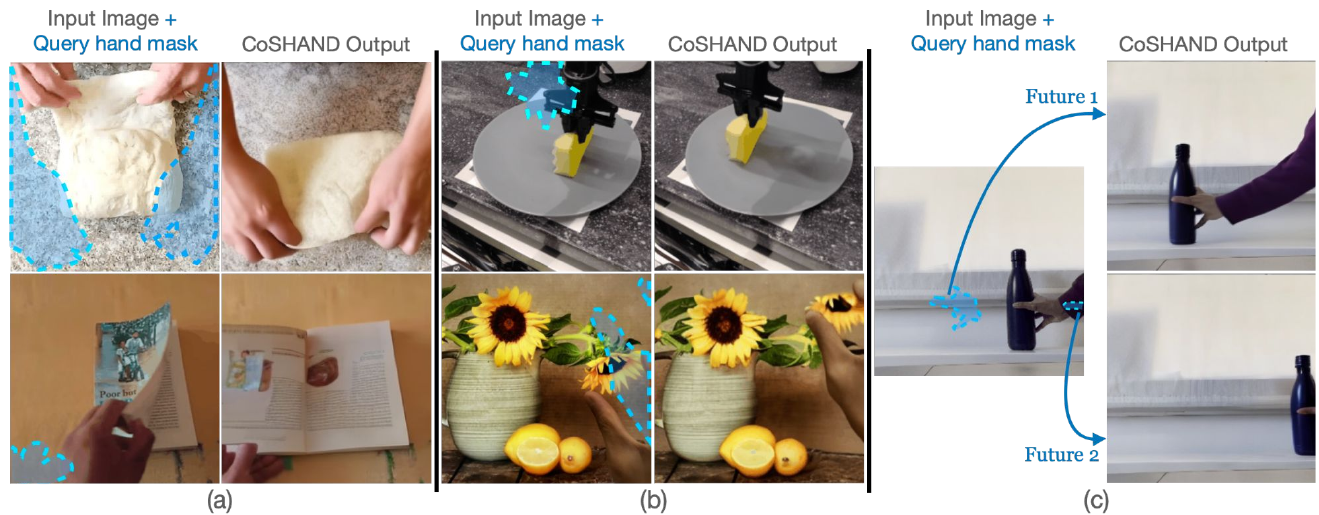}
    \captionof{figure}{\textbf \method \text{ }synthesizes an image of a future after a specific interaction (dotted blue mask) has occurred. In (a) we demonstrate \method's ability to perform complex manipulations on deformable objects such as kneading dough and opening a book. In (b) we show generalization to robot gripper interactions. In (c) we show it is possible to generate diverging futures given the same input context but different hand controls.}
    \label{ref:teaser}
\end{center}
\input{sec/abstract}    
\input{sec/intro}
\input{sec/relwork}
\input{sec/method}
\input{sec/experiments}
\input{sec/discussion}

%
%
\newpage
\bibliographystyle{splncs04}
\bibliography{main}
\newpage
\input{sec/supplemental}

\end{document}

%% file: sec/abstract.tex
\begin{abstract}

Humans naturally build mental models of object interactions and dynamics, allowing them to imagine how their surroundings will change if they take a certain action. While generative models today have shown impressive results on generating/editing images unconditionally or conditioned on text, current methods do not provide the ability to perform object manipulation conditioned on actions, an important tool for world modeling and action planning. Therefore, we propose to learn an action-conditional generative models by learning from unlabeled videos of human hands interacting with objects. The vast quantity of such data on the internet allows for efficient scaling which can enable high-performing action-conditional models. Given an image, and the shape/location of a desired hand interaction, \method, synthesizes an image of a future after the interaction has occurred. Experiments show that the resulting model can predict the effects of hand-object interactions well, with strong generalization particularly to translation, stretching, and squeezing interactions of unseen objects in unseen environments. Further, \method can be sampled many times to predict multiple possible effects, modeling the uncertainty of forces in the interaction/environment. Finally, method generalizes to different embodiments, including non-human hands, i.e. robot hands, suggesting that generative video models can be powerful models for robotics. 

\end{abstract}

%% file: sec/intro.tex
\section{Introduction}
\label{sec:intro}
Humans and animals have an impressive ability to mentally simulate what would happen to an object if they were to interact with it \cite{craik1967nature}. Our ability to reason about our actions and the resulting dynamics of the objects around us has developed from years of interacting with our environment with our hands. Such mental models and forecasting allow us to simulate realistic experiences depending on a specific interactions. Enabling machines to similarly model the future based on interactions can be useful in a variety of tasks such as robotic planning and augmented or virtual reality.

Current generative models are often conditioned on modalities that are easy to acquire at scale, such as text \cite{singer2022makeavideo, yang2023learning, ho2022imagen}. However, these conditioning modalities are not aligned to the action space to enable machines to predict how objects and the environment change and potentially deform in physically plausible ways according to a \textit{specific} interaction. For example, even with a detailed text conditioning of \emph{``squeezing a pillow so that it deforms horizontally''}, the precise direction and distance of the deformation is hard to capture via text (see Fig. \ref{fig:compare}). The key question is, how do we give machines the ability to imagine interactions with their environment? 

In this work, we propose to learn an action-conditional generative model by leveraging large amounts of unlabeled videos of people interacting with objects using their hands. Given an image, and the shape/location of a desired hand interaction, \method, (\textbf{Co}ntrolling the World by \textbf{S}leight of \textbf{Hand}), synthesizes an image of a future after the interaction has occurred.

We use existing off-the-shelf hand segmentation methods to obtain binary \hoee masks which we can then pair with before and after frames of a video. By using automatic segmentation methods and unlabeled videos (180k examples of real-world hand-object interactions from the SomethingSomethingv2 dataset), we can efficiently obtain a large-scale dataset, highlighting the potential to efficiently scale up the approach to achieve stronger performance and better generalization. Furthermore, using binary masks allows the method to be agent-agnostic enabling capabilities for non-human agents. We fine-tune \method from a pre-trained image diffusion model. Image generative models (such as DallE-2 \cite{ramesh2022hierarchical}) have seen billions of images of hand-objects pairs in a variety of configurations, allowing us to leverage it's strong priors about how objects and hands move with respect to each other. Furthermore, the probabilistic generative nature of diffusion models allows \method to model uncertainty in future states due to environment/interaction forces. 

We perform several quantitative and qualitative experiments on our training dataset and an unseen (In-the-wild) test set that we create. \method can predict the effects of hand-object interactions well. A key achievement is the generalization ability to novel objects and scenes not in the training set. Particularly we see strong results on interactions involving translations, stretching, and compressing, where objects are clear, separable from surrounding objects, and in-frame in the input image. The model performs decently on more complex object interactions such as large rotations and precise deformations involving folding. Interestingly, our method can also predict the effects of a robot gripper's interactions, even though it has  not been trained on robot data; a promising direction towards robotic planning. Finally, we benchmark against various unconditional and text-conditional generative models to show the benefits of our proposed approach of controlling by hand.

%% file: sec/relwork.tex
\begin{figure}
    \centering
    \includegraphics[width=0.89\textwidth]{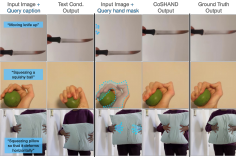}
    \caption{We show that text-conditioning is insufficient to model interactions, whereas hands allow for better control. Columns 1 \& 2 show the input image, query caption, and output of text conditional generation. Columns 3 \& 4 show the input image, query hand mask, and output of \method. Column 5 shows the ground truth output. Notice that \method is able to achieve precise control (including the exact final location of the knife in row 1 and the precise squeezing motion in rows 2 \& 3) which results in a output that is more consistent with the ground truth.}
    \label{fig:compare}
\end{figure}
\section{Related Work}
\label{sec:relwork}

\subsection{Diffusion Models}
Diffusion models have gained tremendous momentum in generative modeling in computer vision. Compared to previous generative architectures such as VAEs and GANs, diffusion models offer many advantages including better training stability, improved coverage of multi-modal distribution of training data, and scalability. Prior works ~\cite{dhariwal2021diffusion} have shown strong experimental results that demonstrates diffusion model beats GAN~\cite{goodfellow2014generative} in image synthesis quality. In DDPM ~\cite{ho2020denoising}, authors propose an effective implementation of diffusion model for image generation by iteratively denoising an image composed of Gaussian noise. Latent Diffusion Model (LDM)~\cite{rombach2022high} proposed an efficient diffusion architecture that performs denoising in latent space of a VAE~\cite{rezende2014stochastic} instead of pixel space.
Due to its fully open-source release, Stable Diffusion has been used as a strong image prior to solve many challenging task in computer vision such as semantic image editing~\cite{brooks2023instructpix2pix,gal2022image,ruiz2023dreambooth}, 3D~\cite{liu2023zero, deitke2023objaverse, wu2023sin3dm}, and segmentation~\cite{amit2021segdiff,ozguroglu2024pix2gestalt}. In this work, we leverage the strong visual priors provided by the internet-pretrained diffusion model to solve the zero-shot hand-conditioned interaction understanding problem.
\subsection{Conditional Generation}
While large-scale diffusion models have recently shown incredible capabilities in generating images and video scenes, methods to control these models are still emerging. Such control is imperative in using these models for extracting plausible world dynamics and making them useful for various applications. Several works have aimed to tackle this control problem from a variety of angles such as text-conditioning by manipulating the weights of the attention maps \cite{brooks2023instructpix2pix, hertz2022prompttoprompt}, camera-view conditioning like in \cite{liu2023zero}, and image based conditioning with a separate set of additional weights like in \cite{hu2021lora} or \cite{zhang2023adding}. Different from a normal text conditioned prediction models, world model~\cite{ha2018world} is a class of models that predict the future conditioned on past observation as well as an action. World models have been applied in robotics for manipulation~\cite{hafner2019dream,dai2023learning,du2023video,yang2023learning}, locomotion~\cite{hafner2019dream,wu2023daydreamer}, and planning ~\cite{yang2024video}. We propose a unique approach of utilizing hand interaction based image conditioning to predict future states by learning from a large-scale dataset of human videos and generalizing to other embodiments. 

\subsection{Hand Environment Interaction}
Hand-environment interaction forms a significant percentage of embodied human experience, making it an important area of study in computer vision and robotics. Prior works have widely explored this topic from various angles. A line of work in computer vision extensively studies the problem of 3D reconstruction of human hands and objects from images~\cite{Shan20,Ye_2023_ICCV,chen2021model,Ye2022WhatsIY, ye2023vhoi}. Another line of work investigates affordances of objects~\cite{pan2017salgan,ye2023affordance,brahmbhatt2019contactdb,liu2022joint,detry2010refining,puhlmann2016compact}. Instead of recognition and reconstruction, many methods propose synthesizing hand-object grasp~\cite{Cutkosky1989OnGC,hsiao2006imitation,liu2019generating,brahmbhatt2019contactgrasp,ye2023affordance}. Other works incorporate physics simulation to generate dynamic grasps for objects that are physically plausible~\cite{christen2022d}. Different from prior work, our proposed method focuses on predicting the changes in object state such as position, appearance, and geometry (deformable objects) caused by hand motions, from a single view.


%% file: sec/method.tex
\section{Method}
\label{sec:method}
We present our method to solve the task of generating an image depicting a future state of the scene after a certain hand interaction has occurred.


\subsection{Learning From Hands at Scale} \label{method:overview}
Recent large-scale models have shown impressive results on tasks such as image generation, visual question answering, etc. These can largely be attributed to the scale of data with which they have been trained on. For example, Stable Diffusion \cite{rombach2022highresolution} was trained on the LAION-5B dataset \cite{schuhmann2022laion5b} which likely includes numerous examples of people interacting with objects in diverse scenarios. We hypothesize that the model understands something about the state of objects based on a certain hand position. Furthermore, objects are often manipulated by hands in the real world, thus Stable Diffusion has likely seen more objects in different states (such as laptops `opened' and `closed') when the hand is present as well. Therefore the key question is how to unlock control to enable manipulation between states (such as opening a laptop). Allowing the model to understand the relationship between hand motions and object states would enable interaction.

We aim to leverage these priors by using Stable Diffusion as a starting point 
and finetuning a large-scale pretrained image-based model on video, as video datasets provide an easy way to extract before and after states of the world to obtain the affects of the applied interaction. Not only does this give us strong priors about how hands and objects interact, but it also allows us to generalize to objects and scenes beyond the data distribution we train on. Furthermore, the priors learned from these hand-object interactions can be transferred to other embodiments such as robot arms as seen in Sec \ref{sec:qualitative}, enabling generalization beyond hands. We emphasize that \method can be easily scaled to larger video datasets because we use off-the shelf models to represent hands, therefore there are no metadata/annotation requirements.

\subsection{Problem Formulation} \label{method:setup}
\begin{figure*}
    \centering
    \includegraphics[width=\linewidth]{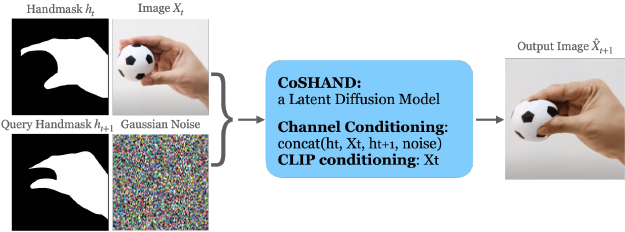}
    \caption{\textbf{\method Method}. We propose a novel approach of controlling by hands to enable manipulating objects in an image. Given an image, the corresponding hand mask, and a query hand mask of the desired interaction, \method synthesizes an image with the interaction applied. Such visual conditioning allows for object interaction.}
    \label{fig:method}
\end{figure*}

We introduce \method, and show an overview in Fig. \ref{fig:method}. Given an RGB image $x_t \in  \mathbb{R}^{H \times W \times 3}$, the corresponding binary hand-mask $h_t \in \mathbb{R}^{H \times W}$ which marks the pixels belonging to the hand in the input image, and a query hand-mask $h_{t+1} \in \mathbb{R}^{H \times W}$ which marks an action taken, our goal is to learn a function f such that 
\begin{align*}
    f(x_t,h_t,h_{t+1})=\hat{x}_{t+1}
\end{align*}
where $\hat{x}_{t+1} \in  \mathbb{R}^{H \times W \times 3}$ is the estimated image and should be perceptually similar to the true but unobserved future $x_{t+1}$. 
\subsection{Conditioning Diffusion Models on Context} \label{method:diffusion}

Image diffusion models have become increasingly popular due to the impressive quality of produced images and the ease and stability of training in comparison to GANs. In DDPM \cite{ho2020denoising}, a model is trained to learn to reverse the diffusion process, which gradually denoises data to produce an image.  In latent diffusion models\cite{rombach2022high}, rather than learning to denoise images in the pixel space, the model is trained to denoise in the latent space of a VAE as this reduces the computational complexity, and increases the expressiveness of the latent code that is learned.

Similar to LDM \cite{rombach2022high}, we use a fixed autoencoder, $\mathcal{E}$, which first encodes an image $x \in \mathbb{R}^{H \times W \times 3}$, into its latent representation $z = \mathcal{E}(x)$. The fixed decoder, $\mathcal{D}$ reconstructs the image from the latent: $\hat{x} = \mathcal{D}(z)=\mathcal{D}(\mathcal{E}(x))$. Both the encoder and decoder are initialized from the pre-trained Stable Diffusion image model \cite{rombach2022highresolution}, allowing us to take advantage of the priors that have been learned. Because we care about modeling an interaction on a \textit{specific} scene, we provide the current image as context to the model. We encode each image $x_{t}$, the corresponding hand masks $h_{t}$, and the future query hand mask $h_{t+1}$, and perform channel-wise concatenation to obtain our full `context' latents $c_i \in \mathbb{R}^{h \times w \times 3c}$. We concatenate the context latents with the latent embedding of the image we are aiming to denoise $z_i \in \mathbb{R}^{h \times w \times c}$ along the channel dimension (where $i$ indicates the diffusion time step and $z_i$ indicates the encoded latent with $i$ steps of added noise).

We learn the diffusion reverse process in this latent space with a U-Net parameterized by $\theta$, $\epsilon_\theta$, trained to iteratively denoise the input image by predicting the noise vector that is subtracted at each time step. We want to model the conditional distribution $p(z|x)$ such that the synthesized image looks semantically similar to the input image. This is because the task entails retaining much of the semantics of the objects and backgrounds in the scene. Therefore, along with channel conditioning, we apply cross-attention to condition the model on a CLIP embedding of the input image $\tau(x_{t})$. We use a frozen CLIP image encoder and randomly drop the conditioning signals $c_i$ and $\tau(x_{t}$) with a probability of 5\% during training. At inference time, we start from Gaussian noise and condition on $c_i$ and $\tau(x_{t})$ using $\epsilon_\theta$. Our learning objective is:
\begin{align*}
   \min_{\theta}\;  \mathbb{E}_{z,c \sim \mathcal{E}(x), i, \epsilon \sim \mathcal{N}(0, 1)}||\epsilon - \epsilon_{\theta}(z_i, c_i, \tau(x_{t}), i))||_2^2.
\end{align*}

%% file: sec/experiments.tex
\begin{figure}
    \centering
    \includegraphics[width=0.91\linewidth]{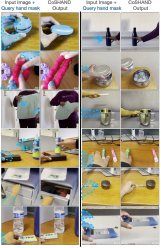}
    \caption{\textbf{Examples In-the-wild (from our lab/home environment)}. We test \method against challenging In-the-wild collected in our home/lab environments. \method remains robust in these scenarios, showcasing its strong generalization ability.}
    \label{fig:realworld}
\end{figure}

\section{Experiments}
\label{sec:experiments}
The central hypothesis of our paper is that by \textbf{(a)} training large-scale diffusion models with our proposed hand conditioning for action control, and \textbf{(b)} learning from a large dataset of unlabeled human videos, we can predict future states of the objects across different scenes and embodiments (human and robot hands). The goal of the experiments section is to test this hypothesis by presenting a set of rigorous and controlled experiments. 


\subsection{Datasets} 
\label{sec:dataset}

\textbf{SomethingSomethingv2 (Human Video Dataset):}
We have seen strong results when training large models with large-scale datasets, as it leads to impressive results in open-world settings with strong performance and generalization. 
To leverage this, we finetune our model on the SomethingSomethingv2 dataset \cite{goyal2017something}.
The dataset contains over 180k videos of humans performing pre-defined, basic actions with everyday objects, for example `moving something up', `putting something next to something', or `pushing something from left to right`. To obtain hand masks, we use Segment Anything \cite{kirillov2023segment} with the provided bounding boxes as prompts.
We split each video into frames at a frame-rate of 12 FPS and sample before and after images at intervals of 3.
At training time, we sample one frame of context and one later target frame.

\textbf{In-the-wild (Self-Recorded Video Dataset):}
We are able to achieve generalization to different environments and objects not present in the original dataset. To showcase examples beyond the test set of the SomethingSomethingv2 dataset, we record 45 videos in our labs/homes of interactions between our hands (single and bi-manual) and a variety of objects in and around our lab/home environments. We aim to find object categories outside the dataset such as dough and white boards (however due to the great diversity of SomethingSomethingv2 object categories do overlap but of course instances are unique). Some samples and results are visualized in our qualitative results (see Fig. \ref{fig:realworld}). 

\textbf{BridgeDatav2 (Robot Dataset):}
Because many high-level object dynamics and interactions are similar across embodiments, we show promising zero-shot (no further training/fine-tuning) generalization results. We use select examples from various environments in the BridgeData V2 \cite{walke2023bridgedata} which contains over 60k trajectories of the WidowX 250 6DOF arm interacting with a variety of objects in 24 different environments. 
\input{tab/baselines_ssv2val}

\subsection{Baselines and Metrics}
\label{sec:baseline}
Central to object deformations and interaction planning is understanding the effects of manipulating of a scene. Therefore, we compare our method of using \hoee conditioning against baselines, which perform image generation with different forms of control: 
\begin{itemize}
    \item \textbf{Masked Conditional Video Diffusion (MCVD)} \cite{voleti2022mcvd}: Prior work has investigated predicting the future conditioned on previous frames. MCVD proposes a masked conditional video diffusion model that can be used for video prediction (forward and backward), unconditional generation, and interpolation. For comparison, we trained MCVD for conditional future prediction.
    \item \textbf{Unconditional Generation (UCG)}: While MCVD is a strong baseline, for a fair architecture comparison we also finetune the pre-trained Stable Diffusion image model with the input frame as conditioning.
    \item \textbf{InstructPix2Pix (IPix2Pix)}: is a strong text editing method. We evaluate InstructPix2Pix on this dataset (with captions provided in the SSv2 metadata) to show the difficulty of text based editing.  
    \item \textbf{Text-Conditional Generation (TCG)}: We also finetune the pre-trained Stable Diffusion model to condition on the input image along with our captions (provided in the SSv2 metadata) fed in through cross-attention layers. This allows for a more fair architectural comparison. 
    \item \textbf{Ours (\method)}: Our method is trained with an image, its corresponding handmask, and a query handmask (indicating the action), and is expected to predict the target image. Note that during evaluation, the mask can be replaced for a robot arm mask as shown later in section \ref{sec:qualitative}. 
\end{itemize}
We evaluate our method and baselines on three metrics that capture various aspects of image similarity, including: PSNR which measures a scaled mean-squared error, SSIM which measures the structural similarity \cite{ssim}, and LPIPS which measures perceptual similarity  \cite{lpips}.

\textbf{Training Details:} We finetuned our model on an 8×A100-80GB machine for 7 days and use AdamW \cite{loshchilov2019decoupled} with a learning rate of $10^{-4}$ for training. We reduce the image size to 256 × 256 (with a 32 × 32 latent dimension), in order to be able to increase the batch size to 192 because larger batch sizes allows for increased training stability and convergence rates. (Note that with more compute and a larger latent space, we could achieve higher quality results as more spatial information would be preserved). We experiment with variations of the classifier free guidance scale at test-time (see \ref{sec:ablations}).  UCG, TCG, and, \method are initialized with the Stable Diffusion provided checkpoint.


\subsection{Quantitative Analysis}
\label{sec:quantitative}
We show numerical results on the test set of SomethingSomethingv2 in Table \ref{tab:ssv2}. We see that \method outperforms all other baselines in all three metrics including PSNR, SSIM, and LPIPS. We notice that MCVD performs poorly compared to other baselines likely due to the lack of pretrained priors or hand controls. UCG performs better however it still lacks hand controls and therefore cannot synthesize specific futures. Text conditioning allows for high-level controls, however most text-conditioning methods are unable to perform drastic object manipulations therefore IPix2Pix and TCG do not show great performance. Finally, \method shows the strongest performance as it uses Stable Diffusion priors and is trained on a large scale manipulation dataset. 

In Table \ref{tab:realworld} we show numerical results on our In-the-wild dataset, showing strong generalization capabilities. \method beats all baselines. Similar trends hold in terms of large image model priors being key for generalizability. Furthermore, we support these metrics with qualitative comparisons as shown in Fig. \ref{fig:compare}. We show that  \method is able to accurately model the manipulation while TCG only vaguely follows the desired interaction.
\begin{figure}
    \centering
    \includegraphics[width=0.899\textwidth]{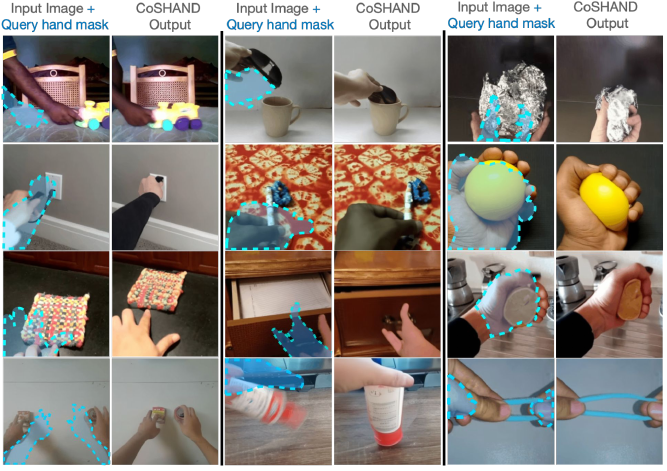}
    \caption{\textbf{We show that \method can perform complex manipulations on a variety of rigid and deformable objects.} We show interactions such as squeezing a lemon, closing a drawer, rotating a bottle, and placing items inside cups, which requires understanding of deformable and articulated objects, as well as occlusion. In columns 1, 3 \& 5 we visualize the input image and the query hand mask of the desired interaction. Columns 2, 4 \& 6 portray the respective outputs of the applied hand interaction.}
    \label{fig:ssv2image}
\end{figure}
\subsection{Qualitative Analysis}
We conduct a qualitative analysis to understand what are success/failure cases of \method and its generalization capabilities. Note that since \method is deterministic, given the same mask and context different samples produce slightly different outputs. Therefore, during our qualitative analysis, we sample the model 4 times and manually asses the outputs and visualize the best of 4. We notice that the variations for most samples are natural and usually correspond to uncertainty of interaction/environment forces.

\label{sec:qualitative}
\textbf{Results on SomethingSomethingv2:} We visualize predictions on the validation set of SomethingSomethingv2 in Fig. \ref{fig:ssv2image}. Due to the diversity in actions that \method was trained on, it is able to understand interactions such as pushing, pulling, stretching, and compressing rigid/deformable objects very well. We also see decent success for more complex interactions such as putting objects inside other objects, opening/closing things, and folding. The training dataset includes bi-manual manipulation data, so we are able to accurately model one and two hand interactions, a strong ability that can aid in downstream robotics tasks. Furthermore, we can model secondary contact forces (e.g. when someone pushes pencil which in-turn pushes a block that it is in contact with, see Fig. \ref{fig:realworld} row 5 column 3/4).

\begin{figure}
    \centering
    \begin{minipage}[t]{0.47\textwidth}
        \centering
        \includegraphics[width=\textwidth]{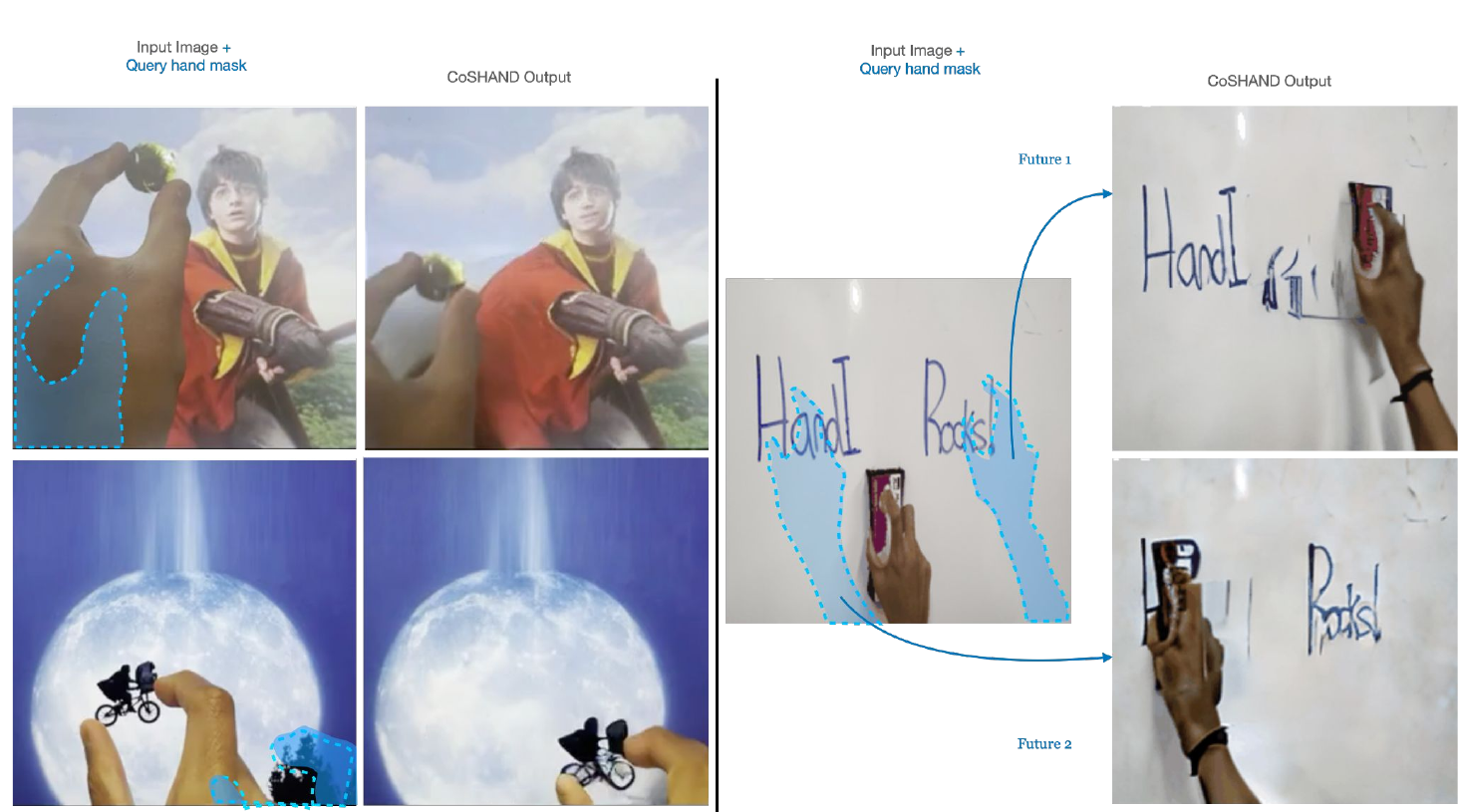}
        \caption{\textbf{\method can edit images and produce different futures}. We can move objects around in famous movie scenes such as the snitch from Harry Potter and the bike from E.T. Furthermore, we show that conditioned on the same input context but different hand mask trajectories, \method predicts an \textbf{alternate future}, while maintaining the photorealism of the predicted future frames.}
        \label{fig:moviesandfutures}
    \end{minipage}
    \hfill
    \begin{minipage}[t]{0.52\textwidth}
        \centering
        \includegraphics[width=0.9\textwidth]{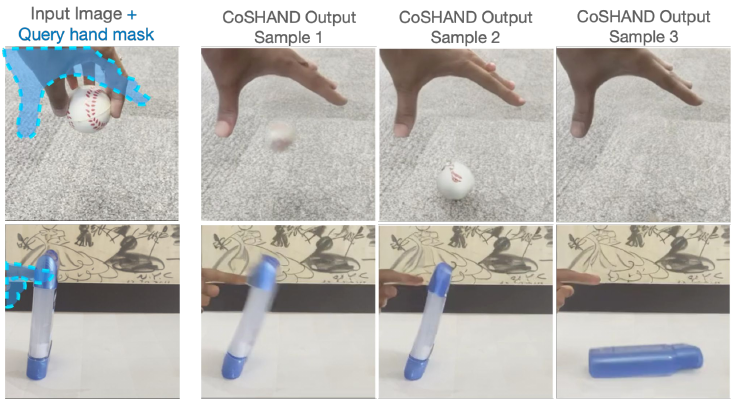}
        \caption{\textbf{Examples where forces are ambiguous}: The force of an interaction or the environment affecting the interaction may be ambiguous and therefore there may be many possible futures. In column 2-4 we show three different samples taken from \method showing the diversity in the outputs when there is uncertainty from interaction.}
        \label{fig:grav}
    \end{minipage}
\end{figure}

\noindent\textbf{In-the-wild results: } We investigate the models ability to generalize to open-world setting. As seen in Fig. \ref{fig:realworld}, the method works on unseen objects/environments likely due to the large diversity in the training dataset. Furthermore, \method is able to model situations where the hand does not remain in contact with the object. In Fig. \ref{fig:grav} the last three columns depict three different samples of \method, reasoning about various possible futures when the forces of the interaction are ambiguous and therefore many futures are possible. 

\noindent\textbf{Predicting two distinct futures: } We show the capabilities of \method to predict two futures given different conditioning. For example, in column 2 of Fig. \ref{ref:teaser}, the model initially sees the hand holding a bottle, but then the person decides to move their hand either right or left, and the model is able to accurately synthesize both futures. Similarly, in Fig. \ref{fig:moviesandfutures} the \method is able to model different motions to erase a whiteboard. Providing a visual cue to this controllable model allows us to detail two possible futures which can be highly useful for example in planning or simulation. 


\textbf{Robustness to Hand Masks:}
While an off-the-shelf Segmentation model can be used to obtain query handmasks as done in this work, a question that could arise is `what if a user were to provide hand-drawn approximate hand masks?'. We explore this questions by comparing segmentation masks obtained via off-the-shelf models (‘SegAny’) to manually-drawn ‘high quality’ query hand-masks and manually-drawn ‘low quality’ query hand-masks (where details of hand/fingers are not captured in mask). To obtain manually-drawn handmasks, we used Keynote’s line-drawing tool. \method is robust to even noisy hand masks. However, \method is better able to reason about the interaction when given more detailed handmasks from ‘SegAny’ \& ‘high qual’, indicating that ‘higher quality’ conditioning produces more fine grained outputs. Therefore, one should strive to select the best practically available hand mask. See figure in Supplemental.

\begin{figure*}
    \centering
    \includegraphics[width=0.85\linewidth]{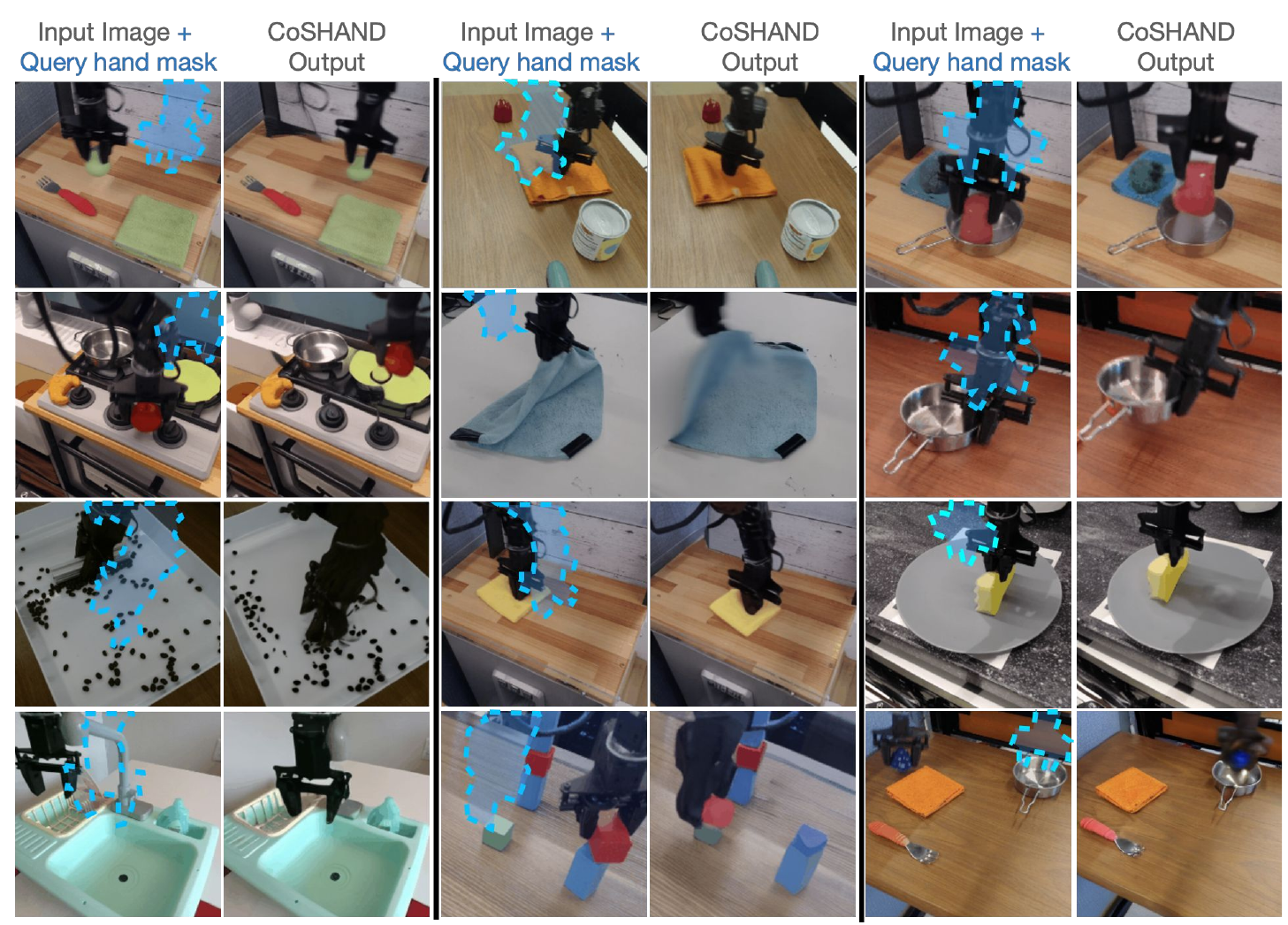}
    \caption{While \method is only trained on hands, it can generalize to robot arms for simple actions. For example, \method can predict reasonably the result of robotic actions including: moving objects around, picking up objects, unfolding cloth, and sweeping granular particles.}
    \label{fig:robotfig}
\end{figure*}
\textbf{Zero-Shot Generalization to Robot Interactions}
There is very little data in robotics and much of the data that exists is constrained to research lab based environments. Therefore, finding ways to transfer knowledge from humans to robots is an important step towards generalizable robotics models. We hypothesize that \method has learned many key aspects of interaction dynamics which though trained on human videos, can be leveraged for other embodiments such as a robot arm. 

We aim to see how \method performs for robot datasets where the interaction is executed by a robot arm. Similar to the setting with hands, we use off-the shelf bounding box and segmentation models to obtain segmentation masks of the robot arm. We input a current frame, the corresponding end effector mask, and the query end effector mask to \method, which then produces the future state of the scene. We evaluate on a small subset of data from BridgeDataV2, and as seen in Fig. \ref{fig:robotfig}, we see good performance on simple object/action pairs such as pushing a towel or picking up and moving a cup. This indicates a promising path towards enabling robots with the ability to build mental models of objects by watching humans interact with the world. 

\noindent\textbf{Image Editing:} Controllable content creation is a key aspect for many designers in the arts and gaming industries, and \method allows for a very natural way to edit scenes through use of the hand. 
To perform this edit, we capture an image of our hand overlaying and manipulating the image as input to \method. In Fig. \ref{ref:teaser}, we show an edit on a painting where the user lifts up a wilting flower. Furthermore, in Fig. \ref{fig:moviesandfutures} we show editing the location of the snitch from Harry Potter, and the bike from the movie E.T. \\ \\  

\subsection{Ablation Study} 
\label{sec:ablations}
We perform several rigorous ablations to test our hypothesis (see Table \ref{tab:ablations} \& Fig. \ref{fig:cfgvs} for quantitative results):
\begin{itemize}
    \item \textbf{No SD prior}: As hypthesized, training \method from scratch without leveraging Stable Diffusion pretraining, results in poor performance, likely due to the lack of prior knowledge of hand-object interactions.
    \item \textbf{No CLIP conditioning}: We remove CLIP conditioning signal and retrain the model to verify CLIP embedding effectiveness. Performance drops on all metrics, likely because without the clip embedding, the overall semantics of the image are lost, making reconstructing details difficult.
    \item \textbf{Less Training Data}: When we train \method with 10\% of the data, performance and generalization suffer. This indicates that as the amount of training data increases, we will see increase in the model's capabilities.
    \item \textbf{Ours + Context}: We hypothesize that providing more frames of context as input to \method would result in better performance as a temporal understanding is induced. We test this hypothesis by training \method with 4 frames of context and their corresponding handmasks, and querying one step into the future. As expected, this improves performance. However, we note that this setting is less common as multiple context frames are often unavailable, therefore we do not delve deeply into this setting.
    \item \textbf{Varying CFG scale}: We notice that increasing the guidance scale beyond a certain threshold, 2.5 in our case, causes the image becomes less dynamic likely due to negative correlation between variance and cfg as seen in Fig. \ref{fig:cfgvs}. This is because a higher cfg causes the model to more closely replicate the input image $x_{t}$. On the other hand, guidance scales less than 2.0 result in the original image being ignored, hence decreasing performance. See Fig. \ref{fig:cfgvs} for visualization of numerical analysis.
\end{itemize}
\input{tab/ablations}



\subsection{Limitations}
\label{sec:limiations}
Our approach aims to learn plausible manipulations between hands and objects, therefore we find our model does not perform well in very unrealistic settings (e.g.: using hands to change shapes of clouds, pushing buildings, etc). Furthermore, we find that when its not obvious if two objects are separated, sometimes interactions with one object will result in ambiguous surrounding objects also being altered. 

%% file: tab/baselines_ssv2val.tex

\begin{table}[ht]
    \begin{minipage}{.5\linewidth}
        \flushright
        \centering
        \footnotesize
        \begin{tabular}{@{}lcccc@{}}
        \toprule
        \multicolumn{1}{l}{Method}  & SSIM $\uparrow$ & PSNR $\uparrow$ & LPIPS $\downarrow$\\
        \midrule
        MCVD & 0.231 & 8.75 & 0.307 \\
        UCG & 0.340 & 12.08 & 0.124 \\
        IPix2Pix & 0.289 & 9.53 & 0.296 \\
        TCG & 0.234 & 9.05 & 0.221 \\
        Ours & \textbf{0.414} &\textbf{ 13.72} & \textbf{0.116} \\
        \bottomrule
        \end{tabular}
        \subcaption{\textbf{Results on validation subset of SomethingSomethingv2.} All metrics demonstrate that our method is able to outperform the baselines by a significant margin.}   \label{tab:ssv2}
    \end{minipage}
    \hspace{\fill}
    \begin{minipage}{.5\linewidth}
        \flushright
        \centering
        \footnotesize
        \begin{tabular}{@{}lcccc@{}}
        \toprule
        \multicolumn{1}{l}{Method}  & SSIM $\uparrow$ & PSNR $\uparrow$ & LPIPS $\downarrow$\\
        \midrule
        MCVD & 0.373 & 11.487 & 0.282  \\
        UCG & 0.458 & 13.858 & 0.210  \\
        IPix2Pix & 0.498 & 13.594 & 0.275  \\
        TCG & 0.454 & 14.201 & 0.207  \\
        Ours & \textbf{0.576} &\textbf{ 18.156} & \textbf{0.125}  \\
        \bottomrule
        \end{tabular}
        \subcaption{\textbf{Results on In-the-wild test set.} We show that our method can generalize to examples not in the training distribution and that our method outperforms baselines.}
    \label{tab:realworld}
    \end{minipage}%
    \vspace{-0.3cm}\caption{\textbf{Quantiative Results on SomethingSomethingv2 and In-the-wild datasets.}}
\end{table}

%% file: tab/ablations.tex
\begin{figure}
    \centering
    \begin{minipage}{.48\textwidth}
        \flushright
        \centering
        \footnotesize
        \begin{tabular}{@{}lcccc@{} }
        \toprule
        \multicolumn{1}{l}{Method}  & SSIM $\uparrow$ & PSNR $\uparrow$ & LPIPS $\downarrow$ \\
        \midrule
        No SD prior & 0.376 & 12.36 & 0.116 \\
        No CLIP Cond & 0.366 & 11.76 & 0.173 \\
        Less Data & 0.369 & 12.45 & 0.127 \\
        Ours & \underline{0.423} & \underline{14.00} & \underline{0.108} \\
        Ours + Context & \textbf{0.448} & \textbf{14.76} & \textbf{0.088} \\
        \bottomrule
        \end{tabular}
        \subcaption{\textbf{Table 9: We perform ablations on our method}. Note that the stable-diffusion priors and size of the dataset contribute to significant performance gains. Furthermore,  when more context is available, the model is able to better reason about the next state (bolded last row).}\label{tab:ablations}
    \end{minipage}%
    \hfill
    \begin{minipage}{0.48\textwidth}
        \centering
        \includegraphics[width=\textwidth]{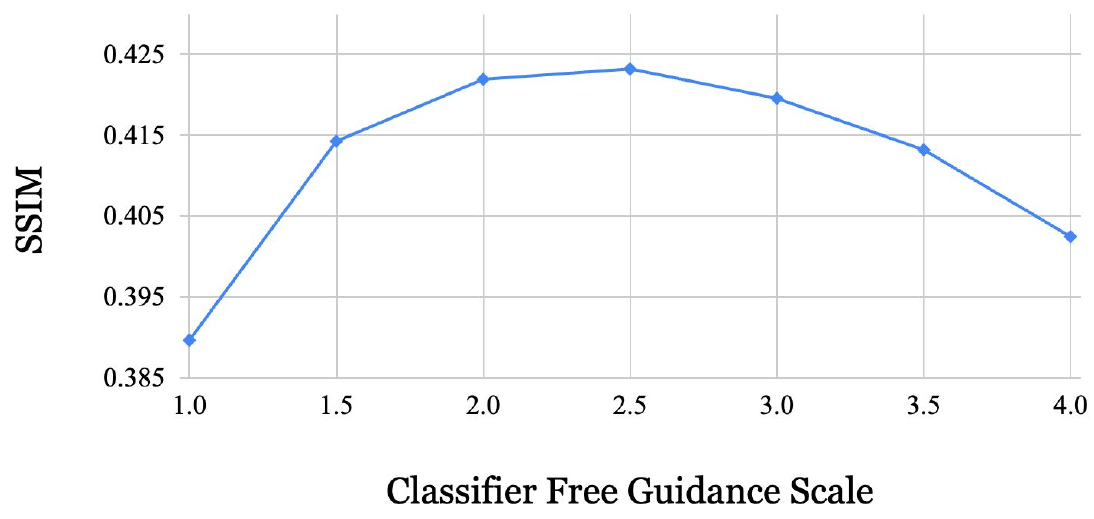}
        \subcaption{\textbf{Fig 9: Classifier-Free Guidance Scale Analysis.} Performance peaks at a cfg value of around 2.5, as too high guidance decreases the variety of the possible generations, while too low of a guidance ignores the input frame. }
        \label{fig:cfgvs}
    \end{minipage}
    \vspace{-0.3cm}

\end{figure}

%% file: sec/discussion.tex
\section{Discussion}
\label{sec:discussion}

In this work, we propose hand conditioning as a new method for interacting with a scene, allowing machines to build a mental model of the future conditioned on a specific action. We introduce \method, a method that leverages large-scale pretrained models and and learns from a large source of unlabeled data to build a representation of interaction.  We show experiments on the Something-Something v2 dataset, real world examples, zero-shot generalization to robot grippers, and interesting image edits. We note that \method does not require any metadata and can therefore be easily scaled to improve results on robot datasets and unrealistic image edits. This opens up exciting applications in robot planning, controllable image generation, and augmented or virtual reality.\\ \\
\textbf{Acknowledgments:} This project is based on research partially supported by NSF \#2202578, NSF \#1925157, and the National AI Institute for Artificial and Natural Intelligence (ARNI). SS is supported by the NSF GRFP.

%% file: sec/supplemental.tex
\section*{Controlling the World by Sleight of Hand: Supplementary Material}
\label{sec:supp}
Please see \url{coshand.cs.columbia.edu} for webpage and code. 

\subsubsection*{A. Dataset: SomethingSomethingv2}

The dataset contains over 180k videos of humans performing pre-defined, basic actions with everyday objects.
The actions fall under 174 prescribed categories.
We exclude videos that belong to the `pretending' category and videos that do not include hands, as these are not relevant to our task.
We utilize hand bounding box annotations provided in the Something-Else dataset \cite{CVPR2020_SomethingElse}, however we note that bounding boxes can quite easily and accurately be obtained from off-the-shelf methods such as GroundingDINO \cite{liu2023grounding} making this approach scalable to larger-scale datasets.

\subsubsection*{B. Dataset: In-the-wild Dataset}

To test the robustness of \method outside of the SomethingSomethingv2 dataset, we collected 45 videos of ourselves interacting with objects in our lab and home settings against various backgrounds. These interactions are listed below (note that some of these actions have multiple videos associated with them, hence only 25 actions are present but with possibly multiple instances per action): opening books, knocking a bottle over, pushing a bottle, lifting a box, opening a container, pulling a chair, pushing a chair, opening a drawer, moving a keyboard across the table, putting pack of gum inside drawer, squishing a soft ball, poking play dough, rotating a bulb, pushing a pencil which then pushes a box, stacking eraser and spray, stacking eraser and marker, separating two markers, moving an eraser across the whiteboard, moving a marker to the side, putting an eraser inside a mug, putting a box inside a mug, moving a pencil behind a mug, separating a picture frame and a bottle , moving a bottle to the side, squishing a pillow. 

\begin{figure}
    \centering
    \includegraphics[width=0.9\textwidth]{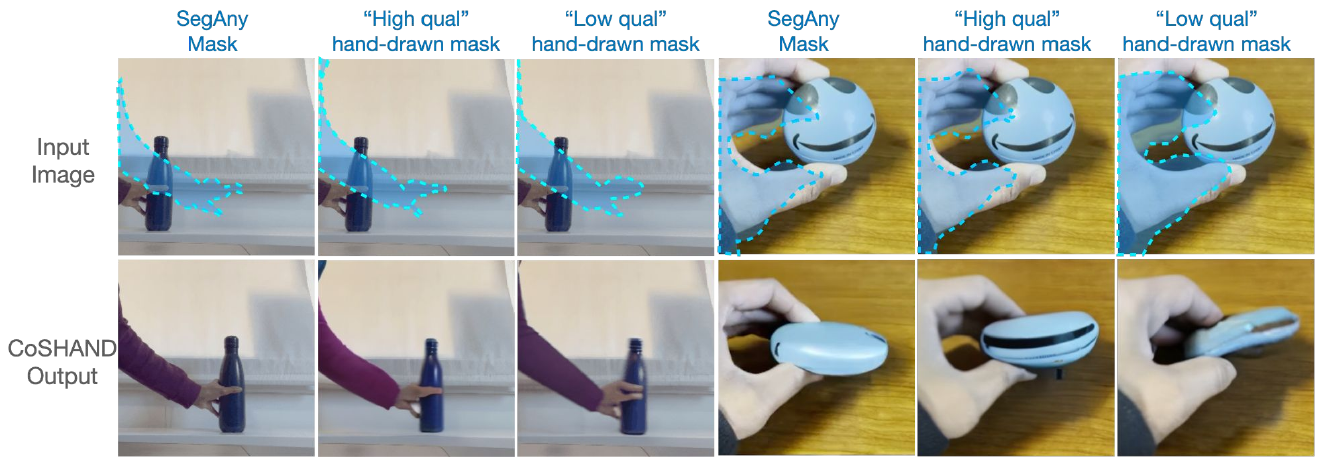}
    \caption{Robustness to handmasks: \method is robust to query handmasks obtained from different sources. Here we show three sources including: off-the-shelf segmentation models (‘SegAny’), manually-drawn ‘high quality’ query hand-masks, and manually-drawn ‘low quality’ query hand-masks (where details of hand/fingers are not captured in mask).}
    \label{fig:handmaskrobustness}
\end{figure}
